\documentclass{elbioimp2}
\usepackage[utf8]{inputenc}

\usepackage[backend=biber,style=vancouver]{biblatex}
\usepackage{csquotes}

\usepackage{amsmath}
\usepackage{amsfonts}
\usepackage{tabularx}
\usepackage{enumitem}
\usepackage{subcaption}

\title{Using reinforcement learning to improve drone-based inference of greenhouse gas fluxes}
\shorttitle{Using reinforcement learning to improve drone-based inference of greenhouse gas fluxes}

\author{Alouette van Hove\affiliation{Department of Geosciences, University of Oslo, Oslo, Norway}\affiliation{dScience - Centre for Computational and Data Science, University of Oslo, Oslo, Norway}\affiliation{E-mail any correspondence to: a.van.hove@geo.uio.no}, Kristoffer Aalstad$^1$ and Norbert Pirk$^1$} 
  
\shortauthor{van Hove et al.}
\elbioimpreceived{12 Oct 2022}  
\elbioimppublished{14 Aug 2023}
\elbioimpfirstpage{1}
\elbioimpvolume{3}
\elbioimpyear{2023}
 
\begin{document}
\maketitle

\begin{abstract}
  
Accurate mapping of greenhouse gas fluxes at the Earth's surface is essential for the validation and calibration of climate models. In this study, we present a framework for surface flux estimation with drones. Our approach uses data assimilation~(DA) to infer fluxes from drone-based observations, and reinforcement learning~(RL) to optimize the drone's sampling strategy. Herein, we demonstrate that a RL-trained drone can quantify a CO$_2$ hotspot more accurately than a drone sampling along a predefined flight path that traverses the emission plume. We find that information-based reward functions can match the performance of an error-based reward function that quantifies the difference between the estimated surface flux and the true value. Reward functions based on information gain and information entropy can motivate actions that increase the drone's confidence in its updated belief, without requiring knowledge of the true surface flux. These findings provide valuable insights for further development of the framework for the mapping of more complex surface flux fields.

\keywords{reinforcement learning, data assimilation, drones, climate science}
\end{abstract}

\section{Introduction}

To date, the inability of climate models to capture the large spatio-temporal variability of land-atmosphere interactions remains a significant source of uncertainty in climate change projections~\cite{crossley_2000, srivastava_2021}. Model validation and calibration is complicated by the scale mismatch between land-atmosphere processes captured by site observations ($\leq1\,\text{km}$), and the grid scales at which climate models are run ($1-100\,\text{km}$). We develop a framework for the mapping of greenhouse gas fluxes at the grid-scale of climate models using drone observations. Our aim is to facilitate model calibration and validation, and ultimately contribute to reducing uncertainties in climate projections.

In the past decade, there has been increased interest in using meteorological sensors on drones to estimate fluxes. For example, \cite{manies_2019}, \cite{allen_development_2019} and \cite{galfalk_sensitive_2021} applied a mass balance approach to quantify the strength of a methane hotspot from drone observations of gas concentrations downwind and upwind of a source. Moreover, \cite{shah_near-field_2019} used drone observations in combination with the inversion of a Gaussian plume model to approximate fluxes. Recently, \cite{pirk_2022} applied data assimilation~(DA) techniques to fuse drone observations with an atmospheric model to estimate surface heat fluxes. We build upon the latter work by taking a similar approach for the source term estimation of greenhouse gas fluxes. 

Optimal sampling strategies for inferring surface fluxes from drone observations are not necessarily evident a priori. On the one hand, the sample duration should be long enough and the number of observations sufficiently large, to capture the time-averaged emission plume as well as possible. On the other hand, sampling for too long a period of time leads to potential errors caused by changes in atmospheric stability, source strength, or prevailing weather during the sampling period. This issue becomes more pressing when sampling larger areas, which is the intention of this framework. This study represents an initial step in the development of a comprehensive framework capable of mapping surface fluxes in diverse environments and under varying atmospheric conditions. We focus on estimating the magnitude of a greenhouse gas surface flux at a known location, assuming static weather conditions.

The strength of reinforcement learning~(RL)~\cite{sutton_reinforcement_2018} is its ability to learn a sequence of actions to complete a task in the most optimal way. RL combined with deep learning  has been successful in source localization tasks using synthetic observations of mobile sensors \cite{wang_2021,loisy_2022,zhao_2022}. Furthermore, \cite{park_source_2022} developed a deep RL-approach to localize and estimate a gas leak with a mobile sensor in a simulated environment. These studies aim to minimize the number of sensor observations to reach the source or achieve a predefined confidence. In contrast, the objective of this study is to develop a sampling strategy with RL that gives the most accurate and precise estimation of a strong CO$_2$ flux while exploiting the full battery life of the drone to collect informative observations. Herein, we use tabular RL to investigate which reward function could be best for this task.

\section{Materials and methods}

\begin{figure}[htp]
  \centering
  \includegraphics[width=0.95\columnwidth]{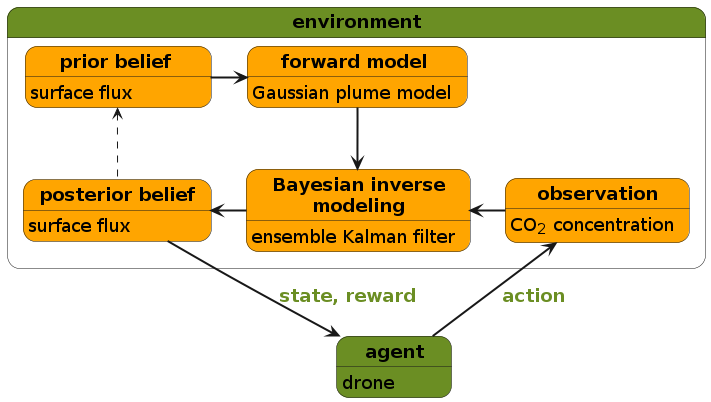}
  \caption{Overview of the framework. Drone observations of gas concentration are fused with a Gaussian plume model. Through data assimilation (in orange) a more accurate estimate of the unobserved surface flux is inferred. A reinforcement learning algorithm (in green) is used to learn an optimal sampling policy for the positions of drone observations to reduce the model uncertainty as much as possible, and consequently increase the accuracy of the estimated surface flux. \label{fig:framework_diagram}}
\end{figure}

Figure~\ref{fig:framework_diagram} illustrates the workflow of our framework. Its components are discussed in the following sections. 

\subsection{Bayesian data assimilation}
Greenhouse gas surface fluxes are inferred from sparse and uncertain drone-based observations of gas concentration. Solving this inverse problem requires a forward (data generating) model that maps from the latent surface fluxes to the drone observations. This is done through $c = \mathcal{F}(\phi) + \epsilon$, where $c$ is an observation of gas concentration, $\mathcal{F}$ is the forward model, $\phi$ is the surface flux, and ${\epsilon}$ is the observation error. We apply Bayesian inference to obtain an estimate of the surface flux.

Following Bayes' theorem: $p(\phi|c)= p(c|\phi) p(\phi) / p(c)$, the prior beliefs about surface flux $p(\phi)$ are updated by a likelihood function $p(c|\phi)$ which measures the likelihood that a flux has generated the observed concentration \cite{mackay2003information}. This update results in posterior belief $p(\phi|c)$, expressing the probability of a surface flux given the observation. In this context, the model evidence $p(c)$ is just a normalizing constant. In the geosciences, we refer to this Bayesian inference exercise as DA \cite{carrassiDataAssimilationGeosciences2018}. 

For the DA scheme, our framework uses an iterative Ensemble Kalman Filter~(EnKF)~\cite{evensen_sequential_1994,Emerick2012} with 100 ensemble members and 4 iterations. In our synthetic experiments, we use the EnKF to estimate surface fluxes with true values between $200\,\text{and}\, 300\,\text{mg}\,\text{CO}_2\,\text{m}^{-2}$\,$\text{s}^{-1}$, corresponding to a strong emission hotspot from e.g. industrial facilities. To ensure strictly positive values, the prior distribution $p(\phi)$ of the surface flux is set to a lognormal distribution with median $100\,\text{mg}\,\text{CO}_2\,\text{m}^{-2}$\,$\text{s}^{-1}$ and mode $30\,\text{mg}\,\text{CO}_2\,\text{m}^{-2}$\,$\text{s}^{-1}$ (shown in Figure~\ref{fig:result_reward}).

\subsection{Gaussian plume model} 
A Gaussian plume model~\cite{brasseur_jacob_2017} is used as the forward model. This time-independent dispersion model assumes that the horizontal and vertical gas concentrations in the emission plume follow a Gaussian spatial distribution. The model's low computational cost enables real-time decision making, which will be essential for the future application of our framework in the field. 

Given this plume model, the steady-state solution for the gas concentration at location $(x,y,z)$ is a function of the form: $c = \mathcal{F}(\phi, U, D, x, y, z, x_{s}, y_{s}, z_{s}, \sigma_{y}, \sigma_{z})$, where $U$ is horizontal wind velocity, $D$ is wind direction, and $(x_{s}, y_{s}, z_{s})$ defines the source location. The standard deviations $\sigma_y$ and $\sigma_z$ of the Gaussian plume are defined in this model by atmospheric stability classes ranging from 1 \textit{"very unstable"} to 6 \textit{"very stable"}~\cite{sigma_gaussian_plume_model}. Herein, all input parameters to the Gaussian plume model, except for surface flux $\phi$, are considered known parameters.

In our experiments, the grid of our two-dimensional horizontal domain is 10 by 10 grid cells with an equal grid spacing of $100\,\text{m}$. The location of the surface flux is $(x_{s},y_{s},z_{s}) = (150, 850, 0)\,\text{m}$. The unperturbed concentration field at $z = 10\,\text{m}$ for a flux of $250\,\text{mg}\,\text{CO}_2\,\text{m}^{-2}$\,$\text{s}^{-1}$ is shown in Figure~\ref{fig:result_reward}. The known parameters are chosen to match a typical field case. Wind speed is $U = 4\,\text{m/s}$ and wind direction is $D = 320^{\circ}$ (north-westerly). The dispersion coefficients $\sigma_y$ and $\sigma_z$ follow from stability class 2 \textit{"moderately unstable"}. Furthermore, a background concentration of $400\,\text{ppm}$ is assumed.

\subsection{Synthetic observations}
The measurement uncertainty of CO$_2$ sensors is typically $\pm30\,\text{ppm}+3\%$ in the measurement range $400 - 10,000~\text{ppm}$~\cite{CO2_sensor}. In the synthetic experiments, drone observations are simulated through $c = \mathcal{F}(\phi) + \epsilon$ with Gaussian error $\epsilon\sim\mathcal{N}(0,\sigma_{obs})$, where $\sigma_{obs}$ is the observation error standard deviation. Assuming a battery life of $32\,\text{min}$ and sampling frequency of $0.1\,\text{Hz}$, we consider a drone taking 12 measurements at each of 16 locations. This is simulated by generating a 2 minute-averaged synthetic observation at each measurement location and for each time step with $\sigma_{obs} = 30/\sqrt{12}\,\text{ppm}$. 

\subsection{Reinforcement learning}
In RL, an agent -~the drone~- receives a reward~$r$ associated with its state~$s_t$ at time $t$ in response to performing an action~$a_t$ (see Figure~\ref{fig:framework_diagram}). In our experimental setup, the action space is defined by 5 possible actions: moving one grid cell in positive $x$-direction, negative $x$-direction, positive $y$-direction, negative $y$-direction, or staying in the current grid cell. Actions that would take the drone off the grid are unavailable. The state space of the drone is defined by its location and the current time step: ($x$, $y$, $t$). Including time in the state space enables the drone to stay in a grid cell for a few steps without getting stuck, and allows for different actions when revisiting grid cells. An episode -~flight~- ends when the drone reaches one of the terminal states: ($x$, $y$, $t=15$).

The agent conducts a trial-and-error search to learn how to map states to actions so as to maximize expected return $G_t$. This is the expected sum of future rewards from time $t$ (assuming discount factor $\gamma = 1$ as is common in episodic tasks). State-action values describe the expected return that follows from taking action~$a$ in state~$s$: $q(s,a) = \mathbb{E} \left[ G_t | s_t=s, a_t=a \right] = \mathbb{E} \left[ \sum_{t=k+1}^{T} r(s_k,a_k) | s_t=s, a_t=a \right]$ where $T$ is the final time step of an episode. The optimal policy -~sampling strategy~- follows from selecting the highest state-action values in all possible states. 

The reward function is the incentive mechanism to teach the drone desirable behaviour. We aim to reward action-selection that updates the initial prior to the episode's final posterior in the DA algorithm as much as possible towards the true, but in practice unknown, flux. The agent receives a reward after each observation. We compare reward functions based on: 
\begin{enumerate}[label=(\alph*)]
    \item The negative of the continuous ranked probability score ($r = -\text{CRPS}$)~\cite{hersbachDecompositionContinuousRanked2000}. The CRPS generalizes the deterministic mean absolute error to probabilistic estimates. In our case it measures the distance between the estimated posterior and the true surface flux. The best possible score of zero only occurs for a degenerate ensemble centered on the truth. This reward function is an incentive for what we want to accomplish: an accurate (low bias) and precise (low variance) posterior flux estimate. As a result, it acts as a benchmark against which we compare the effectiveness of other reward functions.
    Its computation requires knowledge of the true surface flux of the training data. This is not a restriction for offline learning with synthetic data but forms a limitation for further online learning in the field during which the drone could fine-tune its policy based on real-time feedback.
    \item The Kullback-Leibler divergence ($r = D_\text{KL}$)~\cite{mackay2003information, murphy_probabilistic_2023}. The forward $D_\text{KL}(\text{posterior}||\text{prior})$ is a non-negative measure of the amount of information gained when updating the initial prior distribution to the dynamic posterior distribution as a result of the observed data. Knowledge of the true surface flux is not required for this reward function. 
    \item The negative differential entropy of the lognormal posterior ($r = -H$)~\cite{murphy_probabilistic_2022}. The entropy, $H$, is related to the information content of a probability distribution in that it is a measure of uncertainty: It is maximum if the distribution is uniform, and it is zero if the distribution is degenerate. $H$ can be seen as a shifted and scaled version of $D_\text{KL}$ between the estimated dynamic posterior and a uniform distribution~\cite{murphy_probabilistic_2023}. Like $D_\text{KL}$, computing $H$ does not require knowledge of the true flux. 
\end{enumerate}

We estimate the state-action values~$q(s,a)$ using a tabular Q-learning algorithm~\cite{watkins1989}. The learning process is performed over $1.5\times 10^6$ episodes by epsilon-greedy action selection with $\epsilon_{\text{max}} = 1.0$ and $\epsilon_{\text{min}} = 0.01$. The learning rate is set to $\alpha = 0.1$. During the training phase, the environment is reset after each episode of $16$ observations. A true surface flux is randomly selected in the range $200\,\text{to}\,300\,\text{mg}\,\text{CO}_2\,\text{m}^{-2}$\,$\text{s}^{-1}$ such that the resulting state-action value model is trained to find the optimal sampling strategy across a range of fluxes. For each episode, the initial position of the drone is randomly selected among one of the edge grid cells.

\section{Results}

\subsection{RL training convergence}

\begin{figure}[htp]
  \centering
  \includegraphics[width=0.9\columnwidth]{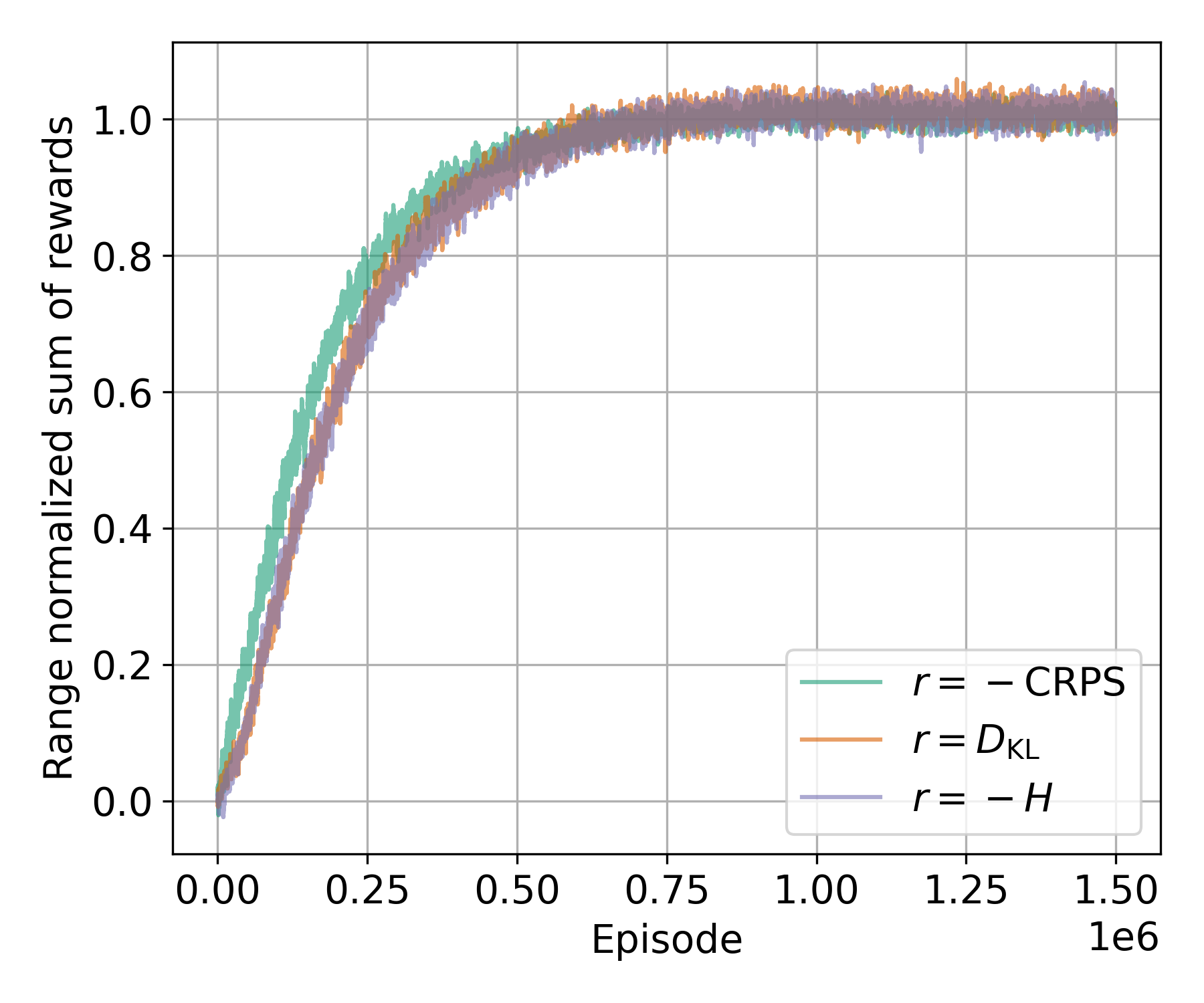}
  \caption{The range normalized sum of rewards per episode for models trained with different reward functions. A moving average over 1,000 episodes is shown. \label{fig:result_learning_curve}}
\end{figure}

Figure~\ref{fig:result_learning_curve} shows the learning curves of the models trained with different reward functions. A moving average over 1,000 episodes is applied to smooth out fluctuations caused by noisy observations and the drone's alternating initial positions. During training, the drone learns to accumulate more rewards per episode. As the exploration rate $\epsilon$ decreases, the learning curves of all models converge. 

\begin{figure*}[h!]
  \centering
  \begin{subfigure}[b]{\textwidth}
        \centering
        \includegraphics[width=0.9\textwidth]{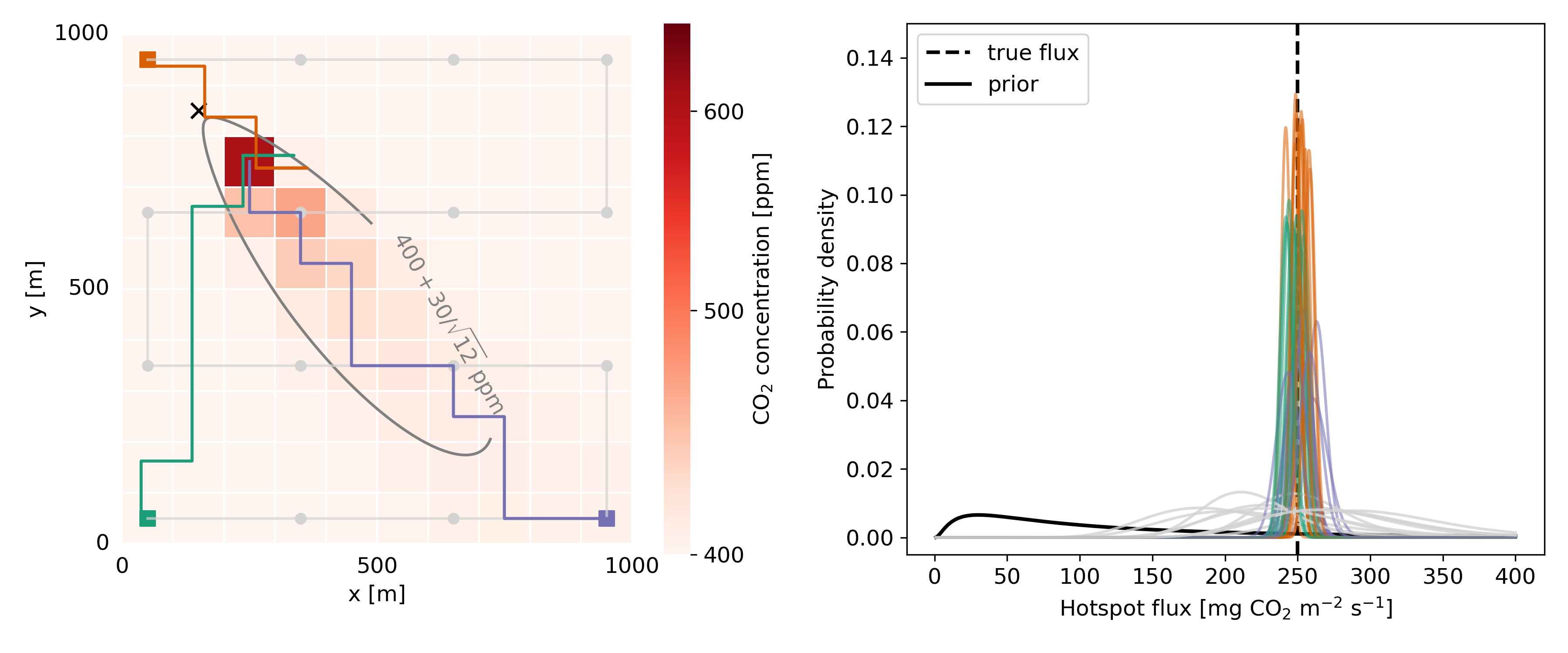}
        \vspace{-3mm}
        \caption{$r = -\text{CRPS}$}
  \end{subfigure}
  \begin{subfigure}[b]{\textwidth}
        \centering
        \includegraphics[width=0.9\textwidth]{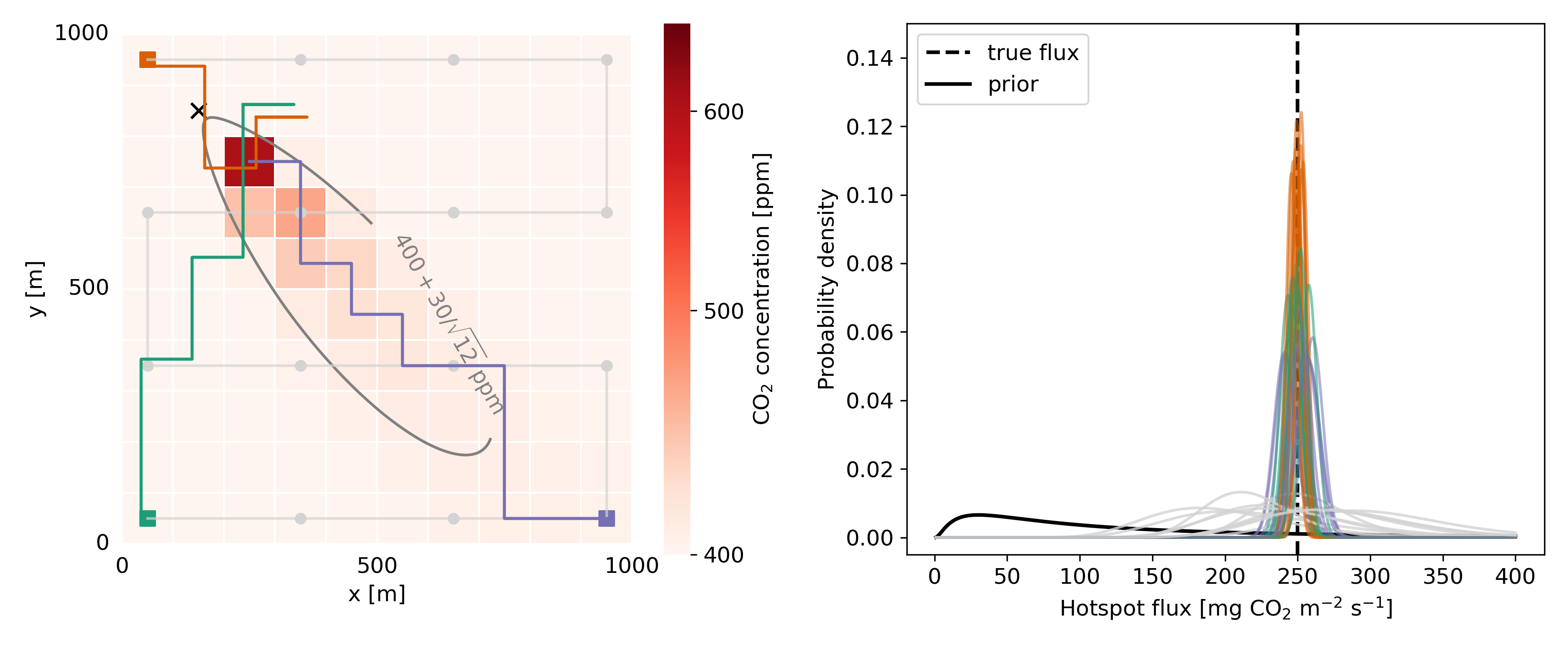}
        \vspace{-3mm}
        \caption{$r = D_{\text{KL}}$}
  \end{subfigure}
  \begin{subfigure}[b]{\textwidth}
        \centering
        \includegraphics[width=0.9\textwidth]{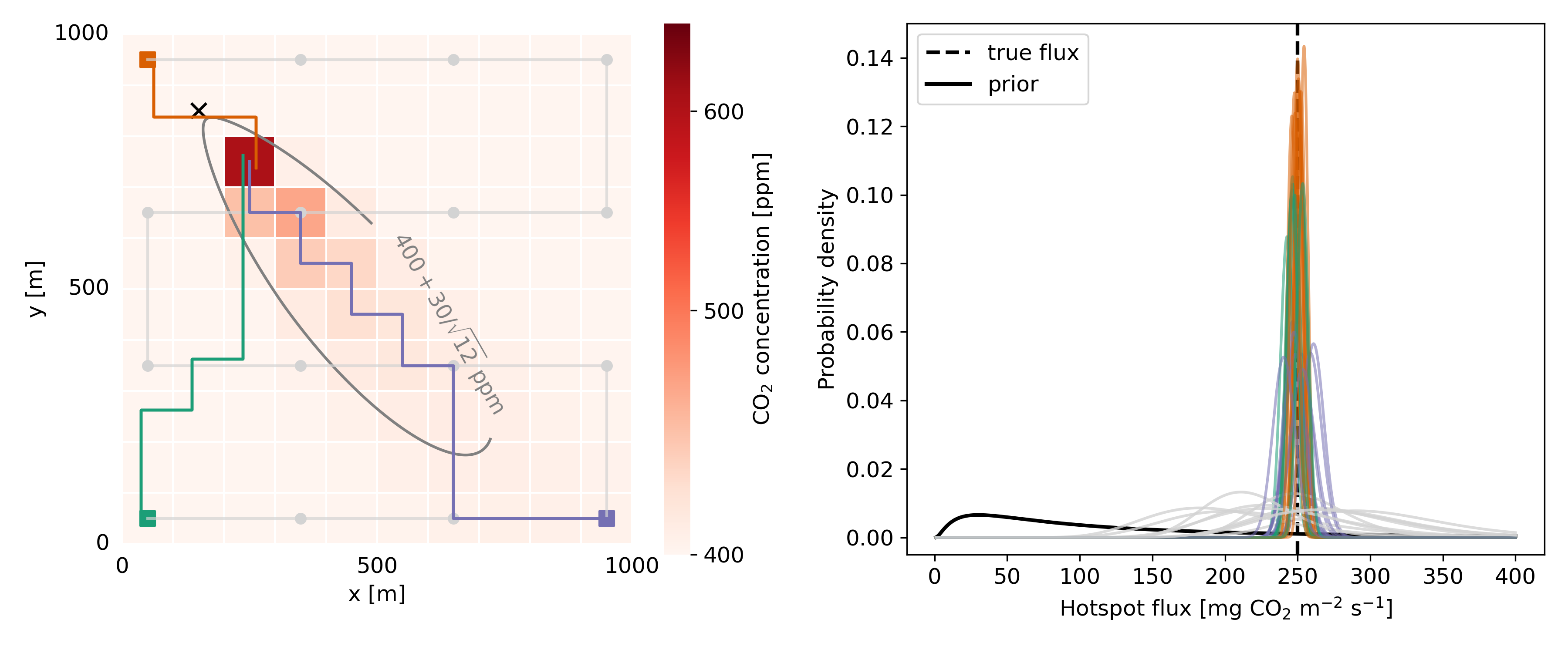}
        \vspace{-3mm}
        \caption{$r = -H$}
  \end{subfigure}
  \caption{Comparison of flux strength estimation by drones trained with different reward functions: (a)~$r = - \text{CRPS}$, (b)~$r=D_{\text{KL}}$ and (c)~$r=-H$, and a drone flying a grid path.
  Left: The unperturbed concentration field at $z = 10\,\text{m}$ for a flux of $250\,\text{mg}\,\text{CO}_2\,\text{m}^{-2}$\,$\text{s}^{-1}$  with $400 + 30/\sqrt{12}\,\text{ppm}$ isoline (dark grey), including the sampling paths of RL-trained drones starting upwind (orange), crosswind (green) and downwind (purple) from the hotspot at $\boldsymbol{\times}$, and a baseline grid path (light grey) with data collection at 16 locations (dots). Right: The prior, true flux and estimated posterior of 10 random individual flights. Differences between flights are a result of noisy observations. \label{fig:result_reward}}
\end{figure*}

\subsection{Sampling strategy}
 
The left panels of Figure~\ref{fig:result_reward} show the sampling strategies of drones trained with different reward functions. Independent from the drone's initial position, the developed optimal sampling strategy is a shortest path to the highest concentration cell. The drone trained with (c)~$r = -H$ stays here for the remaining flight time, and hence takes the maximum possible number of observations in the highest concentration cell. The drones trained with (a)~$r = -\text{CRPS}$ and (b)~$r = D_{\text{KL}}$ mainly sample the highest concentration cell but also observe the adjacent grid cell(s) if the drone started upwind from the hotspot (orange paths) or crosswind from the hotspot (green paths): The drone trained with $r = -\text{CRPS}$ takes one observation in an adjacent grid cell, and the drone trained with $r = D_{\text{KL}}$ takes three observations in adjacent grid cells. In general, the drones starting downwind from the hotspot (purple paths), travel through the emission plume to the highest concentration cell. The drones take the maximum possible number of two samples in this grid cell. 

\subsection{Estimated posterior and final CRPS}

The prior and estimated posteriors after 16 observations are shown in the right panels of Figure~\ref{fig:result_reward}. For comparison, the results of a drone flying a grid path across the emission plume is added. For all RL-trained cases, the prior belief is updated towards the true flux. The estimated posteriors of RL-trained drones are narrower than the estimated posterior of the drone flying a grid path. To evaluate the performance of the synthetic experiments, we use the CRPS of the estimated posterior after the flight. The results of the different models are presented in Table~\ref{tab:CRPS}. Considering the three initial positions investigated in this study, the RL-trained drones have an average $\text{CRPS}$ below $6~\text{mg}\,\text{CO}_2\,\text{m}^{-2}$\,$\text{s}^{-1}$. The drone sampling along a predefined grid path across the plume has an average $\text{CRPS}$ of approximately $ 20~\text{mg}\,\text{CO}_2\,\text{m}^{-2}$\,$\text{s}^{-1}$. 

\begin{table}[htp]
\centering
\begin{tabularx}{0.95\columnwidth} { 
  >{\raggedright\arraybackslash}X
  >{\centering\arraybackslash\hsize=.75\hsize}X 
  >{\centering\arraybackslash\hsize=.75\hsize}X 
  >{\centering\arraybackslash\hsize=.75\hsize}X }
 \hline
 \hline
 {} & upwind & downwind & crosswind \\
 \hline
 $r = -\text{CRPS}$      & {$2.0 \pm 1.4$}  & {$5.5 \pm 4.0$} & {$2.7 \pm 1.9$} \\
 $r = D_{\text{KL}}$  & {$2.1 \pm 1.5$}  & {$5.6 \pm 4.0$} & {$3.1 \pm 2.2$} \\
 $r = - H$               & {$1.9 \pm 1.4$}  & {$5.5 \pm 3.9$} & {$2.5 \pm 1.8$} \\
 \hline
 \hline
grid path & \multicolumn{3}{c}{$20.3 \pm 13.4$} \\
\hline
\hline
\end{tabularx}
\caption{Final posterior CRPS for drones trained with different reward functions starting in locations upwind, downwind and crosswind from a flux of $250\,\text{mg}\,\text{CO}_2\,\text{m}^{-2}$\,$\text{s}^{-1}$, and a drone following a grid path. The average and standard deviation over 5,000 sampling paths is presented. \label{tab:CRPS} }
\end{table}

\section{Discussion}

In this study, we empirically demonstrate the application of RL to develop sampling strategies for drone observations to infer the strength of a greenhouse gas hotspot. Our findings show that RL-trained drones can outperform drones following a predefined grid sampling path across the emission plume. The optimal RL policies identified in this study remain consistent across various true surface flux values and initial drone positions. The learned sampling strategies select a shortest path to the highest concentration cell and sample that grid cell multiple times. This approach is effective because the DA update in this cell may be very informative due to its lower relative measurement uncertainty in comparison to cells with a lower concentration. As a result, the drone can considerably increase its confidence in the updated belief, which is the objective defined in the design of this RL task.

The results of this study offer valuable insights into the use of error-based and information-based reward functions for improving drone-based inference of greenhouse gas fluxes. Using accurate but noisy data, we find that drones trained with information-based reward functions that do not rely on knowledge of the underlying true field of interest can still be effective in finding an optimal sampling policy. When comparing reward functions based on information gain or information entropy, we find that the choice of reference distribution, whether it is the prior at the start of the flight ($r = D_\text{KL}$) or a uniform probability distribution ($r = -H$), has minor impact on the final estimate of the surface flux strength. Both approaches learn a sampling strategy where the highest concentration cell is sampled multiple times. 

The RL-developed sampling strategies in this study may align closely with human intuition for this source term estimation task where many variables are known. However, in more general tasks with numerous unknown parameters it is much less likely that our intuition approaches the optimal sampling strategy. In future work, we aim to adapt the framework to deal with such general settings, enabling its operation in diverse atmospheric conditions and environments. Additionally, our focus will extend to the mapping of more complex surface flux fields, such as hotspots at unknown locations, multiple hotspots, and jointly inferring different types of greenhouse gas fluxes. These future challenges will markedly increase the dimensionality and complexity of the RL task, surpassing the computational capabilities of tabular RL within a reasonable timeframe. Consequently, future studies will use function approximation algorithms, such as neural networks, to fit the state-action model. 

\subsection{Conflict of interest}
Authors state no conflict of interest. 

\subsection{Resources}
Our software implementation can be found at \url{https://github.com/AlouetteUiO/HotSpot_NMI}.

\subsection{Acknowledgement}
This work was supported by the Research Council of Norway (project 301552 "Upscaling hotspots - understanding the variability of critical land-atmosphere fluxes to strengthen climate models (Spot-On)" and project 333232 "Strategies for Circular Agriculture to reduce GHG emissions within and between farming systems across an agro-ecological gradient (CircAgric-GHG)").This work is a contribution to the strategic research initiative LATICE (Faculty of Mathematics and Natural Sciences, University of Oslo, project UiO/GEO103920) as well as the Centre for Computational and Data Science (dScience, University of Oslo).

\nocite{*}
\printbibliography
\end{document}